\documentclass[10pt,twocolumn,letterpaper]{article}

\usepackage{wacv}              

\definecolor{wacvblue}{rgb}{0.21,0.49,0.74}
\usepackage[pagebackref,breaklinks,colorlinks,allcolors=wacvblue]{hyperref}
\usepackage{multirow}

\usepackage{graphicx}
\usepackage{booktabs}

\usepackage{epsfig}
\usepackage{xcolor, soul}
\sethlcolor{red}
\usepackage{boldline}
\usepackage{amsmath}
\usepackage{amssymb}
\usepackage{soul}
\usepackage{url}
\usepackage{booktabs}
\usepackage{comment}
\usepackage{amsthm, nccmath}
\usepackage{mathtools, cases}
\usepackage{color}
\usepackage{lipsum}
\usepackage{float}
\usepackage{multirow}
\usepackage{xcolor}
\usepackage{hhline,colortbl}
\usepackage{ctable}
\usepackage{hyperref}

\usepackage{amssymb}
\usepackage{pifont}
\definecolor{awesomePINK}{rgb}{1.0, 0.13, 0.32}
\definecolor{DarkGreen}{RGB}{1,50,32}
\definecolor{awesomeGRAY}{rgb}{0.5,0.5,0.5}
\definecolor{awesomeYELLOW}{rgb}{0.99, 0.93, 0.0}
\definecolor{TableYELLOW}{rgb}{0.98, 0.91, 0.71}

\definecolor{Gray}{gray}{0.85}
\definecolor{LightCyan}{rgb}{0.88,1,1}

\usepackage{blindtext}
\usepackage{graphicx}
\usepackage{subcaption}
\usepackage[export]{adjustbox}
\usepackage{wrapfig}
\usepackage{adjustbox}

\usepackage{xcolor}
\usepackage{color}

\usepackage[T1]{fontenc}
\usepackage[utf8]{inputenc}

\def\wacvPaperID{3032} 
\def\confName{WACV}
\def\confYear{2026}

\title{\textit{From SAM to DINOv2}: Towards {\color{red}\underline{Di}}stilling {\color{red}\underline{Fo}}undation {\color{red}\underline{M}}odels to Lightweight Baselines for Generalized Polyp Segmentation}

\author{\parbox{16cm}{\centering
    {\large  Shivanshu Agnihotri$^1$, Snehashis Majhi$^2$, Deepak Ranjan Nayak $^1$, Debesh Jha$^3$}\\
    \vspace{0.2cm}
    {\normalsize
    $^1$ Malaviya National Institute of Technology, Jaipur, India \quad
    $^2$ Côte d'Azur University, France \quad
    $^3$ University of South Dakota, USA }}\\
    \vspace{0.2cm}
    \small{\texttt{drnayak.cse@mnit.ac.in}}
    }

\begin{document}
\maketitle
\begin{abstract}

Accurate polyp segmentation during colonoscopy is critical for the early detection of colorectal cancer and still remains challenging due to significant size, shape, and color variations, and the camouflaged nature of polyps. While lightweight baseline models such as U-Net, U-Net++, and PraNet offer advantages in terms of easy deployment and low computational cost, they struggle to deal with the above issues, leading to limited segmentation performance. In contrast, large-scale vision foundation models such as SAM, DINOv2, OneFormer, and Mask2Former have exhibited impressive generalization performance across natural image domains. However, their direct transfer to medical imaging tasks (e.g.,  colonoscopic polyp segmentation) is not straightforward, primarily due to the \textbf{scarcity} of \textbf{large-scale} datasets and lack of domain-specific knowledge. To bridge this gap, we propose a novel distillation framework, \textbf{Polyp-DiFoM}, that transfers the rich representations of foundation models into \textbf{lightweight segmentation baselines}, allowing \textbf{efficient and accurate} deployment in clinical settings. 
In particular, we infuse semantic priors from the foundation models into canonical architectures such as U-Net and U-Net++ and further perform frequency domain encoding for enhanced distillation, corroborating their generalization capability. Extensive experiments are performed across five benchmark datasets, such as \textbf{Kvasir-SEG}, \textbf{CVC-ClinicDB}, \textbf{ETIS}, \textbf{ColonDB}, and \textbf{CVC-300}. Notably, Polyp-DiFoM consistently outperforms respective baseline models significantly, as well as the state-of-the-art model, with nearly $9 \times$ reduced computation overhead. The code is available at \href{https://github.com/lostinrepo/PolypDiFoM}{GitHub repository}.


\end{abstract}
    
\section{Introduction}
Colorectal Cancer (CRC) remains a serious health challenge worldwide due to its high mortality rates and is ranked third among all cancer types. The polyps, small growths of abnormal cells, in the inner lining of the colon often develop into CRC; hence, their early detection and timely intervention are critical for preventing its progression \cite{morgan2023global}. While colonoscopy remains the gold standard for polyp detection and resection, the manual nature of this procedure makes it laborious, highly observer-dependent, and subject to inter-observer variability. Moreover, studies have reported a substantial polyp miss rate of 6–27\% during colonoscopy \cite{ahn2012miss}, underscoring the need for precise and reliable computer-aided polyp segmentation methods to assist clinicians and improve diagnostic outcomes.

In recent years, several encoder–decoder based CNN architectures have been proposed for polyp segmentation. Early approaches, including U-Net \cite{ronneberger2015u} and its variants, have been widely adopted but demonstrate limited ability to capture fine boundary details. 
Later, more specialized segmentation methods such as PraNet \cite{fan2020pranet}, MSNet \cite{zhao2021automatic}, SFA \cite{fang2019selective}, and M$^2$SNet \cite{zhao2023m} have been introduced to preserve boundary information while capturing the inherent scale diversity of polyps. Despite considerable performance improvements, these models struggle to learn global contextual representations, leading to suboptimal segmentation accuracy and limited generalizability.

Recently, Vision Transformer (ViT)-based models \cite{han2022survey, liu2021swin, wang2021pyramid, tu2022maxvit} have demonstrated significant potential in medical image segmentation tasks by effectively modeling global feature dependencies using self-attention (SA) mechanisms. 
Drawing inspiration from their notable success, a few ViT-based architectures such as CTNet \cite{xiao2024ctnet}, PVT-Cascade \cite{rahman2023medical}, and Polyp-PVT \cite{Dong2023} have been developed recently and have exhibited impressive  polyp segmentation results. However, these models demand relatively higher computational resources and have shown suboptimal generalization performance. Despite notable strides, polyp segmentation during colonoscopy remains difficult primarily due to the large variability in polyp shape and size, the presence of unclear boundaries, and the camouflaged appearance of polyps.

Foundation Models (FMs) such as Segment Anything Model (SAM) \cite{kirillov2023segment}, DINOv2 \cite{oquab2024dinov}, OneFormer \cite{jain2023oneformer}, CLIP~\cite{radford2021learning}, MaskFormer \cite{cheng2021per} and Mask2Former \cite{cheng2022masked} have redefined segmentation pipelines by learning intricate visual patterns and offering strong cross-domain generalization. However, their direct transfer to medical segmentation tasks (e.g.,  colonoscopic polyp segmentation) is challenging because of the scarcity of large-scale datasets and the lack of domain-specific knowledge. In addition, foundation models are highly resource-intensive, demanding substantial parameters and memory, which constrains their applicability in resource-limited clinical settings. This underscores the need for lightweight yet effective solutions while preserving the promising potential of FMs.

To address this, we propose \textbf{Polyp-DiFoM}, a novel distillation-based framework that transfers enriched representations from multiple FMs into lightweight segmentation baselines. Our approach is modular and clinically deployable, designed to enhance the effectiveness and generalization of U-Net and U-Net++ without incurring the computational burden of full-scale FM inference. Specifically, Polyp-DiFoM infuses semantic priors from FMs into the baseline encoder and further refines them via frequency-domain encoding, capturing both low- and high-frequency components to preserve intricate structural details. The framework comprises three core components: a \textbf{Baseline Encoder}, a \textbf{Semantic High-Low Distillation Module}, and a \textbf{Foundational Feature Aware Decoder}, each operating at distinct spatial and semantic information. We evaluate Polyp-DiFoM across five benchmark datasets—\textbf{Kvasir-SEG}, \textbf{CVC-ClinicDB}, \textbf{ETIS}, \textbf{ColonDB}, and \textbf{CVC-300}—and demonstrate that our distilled models consistently outperform their vanilla counterparts, achieving superior segmentation accuracy and generalization under diverse imaging conditions.

In summary, our contributions are three-folds:
\begin{itemize}
    \item \textbf{Polyp-DiFoM:} We introduce a modular distillation framework that injects rich priors from multiple vision foundation models (SAM, DINOv2, CLIP, OneFormer, Mask2Former) into lightweight baselines (U-Net/U-Net++), tailored for clinical deployment.
    
    \item \textbf{Frequency-Aware Distillation:} We analyze FM features in the frequency domain, distilling both low- and high-frequency components to sharpen structural detail and enrich semantic information.
    
    \item \textbf{Generalization Boost:} Polyp-DiFoM consistently outperforms vanilla baselines across five benchmarks with stronger cross-dataset generalization.
    
    \item \textbf{Clinical Efficiency:} Polyp-DiFoM training scheme leverages foundational knowledge with considerably faster inference time, accurate prediction, and is thereby highly suitable for real-world clinical use.
\end{itemize}

\section{Related Work}
\textbf{CNN-based Methods:} 
The widely adopted U-Net \cite{ronneberger2015u} architecture has demonstrated strong performance across a wide range of tasks, including polyp segmentation. Following this, U-Net++ \cite{zhou2018unet++} has been introduced with nested dense connections to improve gradient flow further and refine segmentation results. Later, a plethora of methods have been proposed to address the challenge of weak and irregular polyp boundaries. For example, SFANet \cite{fang2019selective} incorporates an auxiliary decoder with a boundary-sensitive loss, while PraNet \cite{fan2020pranet} leverages reverse attention and parallel partial decoders to progressively highlight complex polyp regions. Another CNN-based model, MSNet \cite{zhao2021automatic}, introduces a multi-scale subtraction module and concatenates multiple such modules pyramidally to improve multi-scale feature representations, allowing the model to effectively deal with polyp scale diversities. Subsequently, a variant of MSNet, called M$^2$SNet \cite{zhao2023m}, has also been developed, employing intra- and inter-layer multi-scale subtraction units for efficient polyp segmentation. Zhou et al. \cite{zhou2023cross} introduces CFA-Net, which combines boundary-aware features with cross-level feature fusion to tackle scale variations and ambiguous polyp boundaries. MEGANet \cite{bui2024meganet} incorporates multi-scale edge-guided attention modules between encoder and decoder to preserve edge information, enhancing segmentation of polyps with weak boundaries, while EMCADNet \cite{rahman2024emcad} introduces efficient multi-scale cascaded decoders for refining segmentation details with fewer parameters. While these models achieve performance improvements, their limited ability to capture long-range dependencies still hinders optimal performance.

\textbf{Transformer-based Methods:} These segmentation models have recently garnered interest from the vision community due to their ability to capture long-range feature dependencies. To this end, a few methods have been developed for polyp segmentation. For instance, in \cite{rahman2023medical}, PVT-Cascade has been introduced to enhance the modeling of global and local contexts through a hierarchical cascaded attention-based decoder. 
Dong et al. \cite{Dong2023} proposes Polyp-PVT, a transformer backbone combined with fusion and identification modules, to extract more powerful and robust features and effectively segment polyps with diverse natures. Recently, Xiao et al. \cite{xiao2024ctnet} has developed CTNet, a contrastive transformer network, that better captures long-range dependencies and provides highly structured feature maps, leading to improved generalization performance. Despite remarkable performance, the intensive resource requirements of these models hinder their practical deployment in resource-limited clinical settings. Further, the generalization performance on unseen datasets remains suboptimal.

\begin{figure*}[t]
  \centering

   \includegraphics[width=\linewidth]{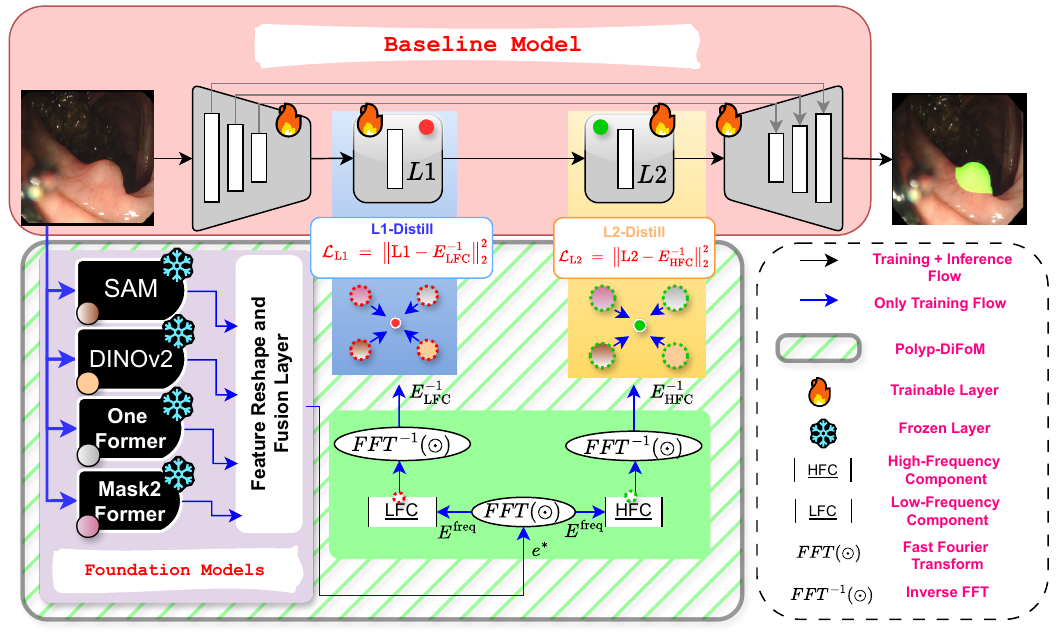}
   \vspace{-0.5cm}

   \caption{\textbf{Overview of the proposed Polyp-DiFoM} framework. Our modular distillation-based architecture transfers rich structural and semantic priors from foundation models (SAM, DINOv2, OneFormer, Mask2Former) into a lightweight segmentation baseline (U-Net).} 


   \label{fig_model}
   \vspace{-0.5cm}
\end{figure*}

\section{Preliminaries and Motivation}

Recent advances of vision foundation models (FMs) such as CLIP~\cite{radford2021learning},  SAM \cite{kirillov2023segment}, DINOv2 \cite{oquab2024dinov}, OneFormer \cite{jain2023oneformer}, and Mask2Former \cite{cheng2022masked} have redefined segmentation paradigms by offering robust, general-purpose representations learned from massive-scale data.
These models exhibit strong generalization across domains, inspiring their adaptation for medical imaging tasks. However, their direct deployment in clinical setting, especially for polyp segmentation—remains constrained by several factors: \textbf{(i) domain gaps} between natural and endoscopic imagery, \textbf{(ii) data scarcity} that exacerbates overfitting, \textbf{(iii) high computational cost} incompatible with real-time use, and \textbf{(iv) semantic granularity mismatches}, as medical segmentation demands pixel-level boundary precision beyond object-level abstraction.

SAM-Mamba~\cite{dutta2025sam} pushes generalization further by combining SAM with adapters and efficient sequence modeling. However, its $103$M trainable parameters and $423$ GFLOPs make it computationally expensive, restricting its use in real-time clinical settings.

In contrast, lightweight architectures like U-Net remain the backbone of medical segmentation due to their encoder–decoder design and skip connections. They offer efficiency and interpretability but often struggle with robustness under inter-patient variability, low-contrast polyps, and cross-dataset generalization. This trade-off between generalization and efficiency raises a critical question: \textit{can such a lightweight framework be scaled to integrate knowledge from multiple foundation models simultaneously, and what strategies can be employed to mitigate the risk of noisy or redundant information that might degrade the performance?} 

\section{Proposed Polyp-DiFoM}
In this section, we present the \textbf{Polyp-DiFoM} framework, a novel and modular distillation-based framework that leverages powerful foundation models (e.g., SAM, DINOv2, OneFormer, Mask2Former, etc.) to enhance the effectiveness and generalization of lightweight segmentation backbones such as U-Net and U-Net++. The methodology, illustrated in Figure~\ref{fig_model}, is designed to transfer rich structural and semantic priors from foundation models into compact baselines, thereby improving polyp segmentation performance while maintaining efficiency. The framework comprises three core components:\textbf{ a Baseline Encoder, a High-Low Distillation Module, and a Foundational Feature Aware Decoder}. Each component is designed to operate on specific spatial and semantic scales, enabling robust segmentation across diverse polyp morphologies and imaging conditions.

\subsection{Baseline Model} We redesign the standard U-Net encoder to produce two complementary latent vectors that capture distinct aspects of the input image: one for global semantic context and another for local structural detail. This dual representation enhances downstream segmentation by enabling richer decoding pathways. Let the input image be a raw endoscopic image $I \in \mathbb{R}^{H \times W \times 3}$, where $H$ = $W$ = $352$. The image is normalized and passed through a series of convolutional blocks:
$I \rightarrow \{f_1^{e}, f_2^{e}, f_3^{e}, f_4^{e}\}, \quad f_i^e \in \mathbb{R}^{\frac{H}{2^i} \times \frac{W}{2^i} \times C_i}$
with $C_i \in \{64, 128, 256, 512\}$ representing the channel dimensions at each stage. For the \textbf{Latent Vector Extraction}, at the bottleneck layer (i.e., after the final encoder block), we extract two latent vectors: \textbf{$L1$} (captures holistic scene-level semantics) and \textbf{ $L2$} (encodes fine-grained spatial and boundary details). The $L1 \in \mathbb{R}^{22 \times 22 \times 256 }$ is obtained via $1 \times 1$ convolutional layer over the deepest feature map $f_4$, while $L2 \in \mathbb{R}^{22 \times 22 \times 256}$ is derived via another convolutional operation over $L1$. These features serve as the student representations to be aligned with the distilled outputs from foundation models.

 \subsection{Semantic High-Low Distillation}
To effectively transfer knowledge from pretrained foundation models (SAM, DINOv2, OneFormer, and Mask2Former) into our BaselineModel, we introduce a two-stage \textbf{Semantic High--Low Distillation} strategy. This approach captures both global semantic richness and fine-grained structural cues, while suppressing redundant or noisy information through frequency decomposition.

\paragraph{Rich Semantic Feature Extraction}

We begin by extracting high-dimensional semantic embeddings $E \in \mathbb{R}^{H \times W \times D}$ from the foundation models, where \( (H,W)  \) is the spatial resolution of feature map, and \( D \) represents the number of channels depending on the model (SAM, DINOv2, Oneformer and Mask2Former). These embeddings are first resized to a common spatial resolution \( H' \times W' \) (typically \( H' = W' = 32 \)) using bilinear interpolation, and then concatenated along the channel dimension to form a unified semantic representation $e^* \in \mathbb{R}^{H' \times W' \times D^*}$. 
\begin{equation}
    D^* = \text{Concat}(D_{\text{SAM}}, \textcolor{blue}{}D_{\text{DINOv2}}, D_{\text{OneFormer}}, D_{\text{Mask2Former}})
\end{equation}

\paragraph{Frequency-Domain Encoding}

To suppress irrelevant information and enrich feature representations, we compute the frequency-domain encoding by applying a 2D Fast Fourier Transform (FFT) \cite{katznelson2004introduction} to the unified semantic representation, which can be expressed as
\begin{equation}
E^{\text{freq}} = \text{FFT}(e^*) \in \mathbb{R}^{H' \times W' \times D^*}
\end{equation}
\begin{equation}
    \text{FFT}(e^*) = \tilde{E}_{\text{r}}(p,q) = 
    \sum_{a=0}^{H'-1} \sum_{b=0}^{W'-1} 
    e^{*}_{r}(a,b)\, e^{-j2\pi\left(\frac{pa}{H'} + \frac{qb}{W'}\right)} 
\end{equation}
Here, \( e^*_{r}  \) denotes the $r^{th}$ channel of $e^*$, and $E^{\text{freq}}$ represents the frequency-domain representation.

We then apply radial frequency masks \( M_{\text{HFC}}, M_{\text{LFC}} \in \{0,1\}^{H' \times W'} \) to isolate the High-Frequency Component (HFC) and Low-Frequency Component (LFC). Further, 2D Inverse FFT is applied to generate final distillation ready components as: 

\[
E^{-1}_{\text{HFC}} = \text{IFFT}(M_{\text{HFC}} \cdot E^{\text{freq}}), \quad E^{-1}_{\text{LFC}} = \text{IFFT}(M_{\text{LFC}} \cdot E^{\text{freq}})
\]
where, inverse FFT can be mathematically presented as:
\begin{align}
\text{IFFT}(Z^{\text{freq}}) & = \tilde{e_r}(a,b) \\
& = \frac{1}{H'W'} 
\sum_{p=0}^{H'-1} \sum_{q=0}^{W'-1} 
Z^{\text{freq}}_r (p,q) e^{j 2\pi \left(\frac{pa}{H'} + \frac{qb}{W'}\right)}
\end{align}
Here, $Z^{\text{freq}}= M_{\text{HFC}} \cdot E^{\text{freq}}=M_{\text{LFC}} \cdot E^{\text{freq}}$ is the frequency-domain input, and $\tilde{e}(a,b)$ is the reconstructed spatial feature representations.

These components are then injected into the baseline model at different stages to guide feature learning. The HFC emphasizes edge and boundary details, while LFC captures global shape and contextual semantics.

\paragraph{L1 Distillation: Semantic-Level Alignment}

The L1 distillation targets the deepest encoder features of the baseline model. We project the teacher’s semantic features \( E^{-1}_{\text{LFC}} \in \mathbb{R}^{32 \times 32 \times D^*} \) to match the student’s feature dimension via \texttt{bilinear interpolation}, and apply then Mean Squared Error (MSE) loss:

\begin{equation}
\mathcal{L}_{\text{L1}} = \frac{1}{HWD^*} \sum_{i,j,k} \left\| L1(i,j,k) - E^{-1}_{\text{LFC}}(i,j,k) \right\|_2^2
\end{equation}

This encourages the student to replicate the semantic abstraction of the teacher, improving contextual understanding and object localization.

\paragraph{L2 Distillation: Structural-Level Alignment}

The L2 distillation further refines the semantically-aligned features obtained from L1 by emphasizing high-frequency structural details. We align them with the teacher’s semantic features \( E^{-1}_{\text{HFC}} \in \mathbb{R}^{32 \times 32 \times D^*} \) using another \texttt{bilinear interpolation}, and compute the MSE loss:

\begin{equation}
\mathcal{L}_{\text{L2}} = \frac{1}{HWD^*} \sum_{i,j,k} \left\| L2(i,j,k) - E^{-1}_{\text{HFC}}(i,j,k) \right\|_2^2
\end{equation}

This alignment sharpens the student’s ability to capture structural boundaries and subtle texture variations—critical for accurate polyp segmentation.

\paragraph{Total Distillation Objective}

The final distillation loss is a weighted combination of both stages:

\begin{equation}
\mathcal{L}_{\text{distill}} = \alpha_1 \mathcal{L}_{\text{L1}} + \alpha_2 \mathcal{L}_{\text{L2}}
\end{equation}

where \( \alpha_1  \) and \( \alpha_2 \) control the relative importance of semantic and structural alignment. In practice, we set \( \alpha_1 = 1.0 \), \( \alpha_2 = 0.5 \) to prioritize semantic abstraction while preserving structural fidelity.

\subsection{Foundational Feature Aware Decoder}
 The decoder takes over as the semantic sculptor—transforming distilled representations into fine-grained polyp segmentation masks. Unlike conventional U-Net decoders that rely solely on spatial upsampling, our decoder is designed to infuse critical polyp-relevant features from foundations model with hierarchical encoder outputs, enabling both boundary sharpness and contextual coherence. The decoder receives two streams of information: \textbf{(i)} the encoder feature maps $\{f_1^e, f_2^e, f_3^e, f_4^e\}$, each of shape $f^e_i \in $ $\mathbb{R}^{\frac{H}{2^i} \times \frac{W}{2^i} \times C_i}$ and \textbf{(ii)} the distilled semantic components $E^{-1}_{\text{LFC}}$ and $E^{-1}_{\text{HFC}}$ to $L1$ and $L2$ latent vectors. These streams of information are concatenated channel-wise and passed through a series of convolutional blocks. By this, the decoder acts as a polyp-relevant, rich semantic translator, converting distilled knowledge into pixel-level segmentation precision.

 \subsection{Multi-Objective Training Strategy}
To optimize both representation learning and knowledge transfer, we adopt a phased multi-objective training protocol that progressively refines the model’s capacity for generalized polyp segmentation. 
\textbf{Phase I: Baseline Initialization}  
We begin by training a vanilla U-Net architecture from scratch. This phase allows the encoder–decoder backbone to establish a stable inductive bias grounded in spatial features, without external guidance from foundation models. Here, the entire model is optimized only with the segmentation loss, which combines Binary Cross-Entropy (BCE) and Dice loss. The loss function for phase 1 is given by:
\begin{equation}
\mathcal{L}_{\text{phase1}} = \mathcal{L}_{\text{dice}} + \mathcal{L}_{\text{bce}} 
\end{equation}

\textbf{Phase II: Joint Distillation}  
In the second phase, we activate the distillation modules and jointly train the U-Net alongside semantic and structural supervision from pretrained foundation models. The model is optimized using both segmentation loss and distillation losses ($\mathcal{L}_{\text{L1}}$, $\mathcal{L}_{\text{L2}}$), enabling the network to absorb rich priors while continuing to refine its spatial representations. 
The loss for this phase is given by:
\begin{equation}
\mathcal{L}_{\text{phase2}} = \lambda_1 \cdot \left( \mathcal{L}_{\text{dice}} + \mathcal{L}_{\text{bce}} \right) + \lambda_2 \cdot \mathcal{L}_{\text{L1}}  + \lambda_3 \cdot \mathcal{L}_{\text{L2}} 
\end{equation}
where $\lambda_1$, $\lambda_2$ and $\lambda_3$ balance the contribution of different loss components. We set $\lambda_1$ =
0.6, $\lambda_2$ = 0.1 and $\lambda_3$ = 0.1 to emphasize semantic learning through segmentation loss, while still encouraging structural fidelity via distillation.

\textbf{Phase III: Polyp Focused Distillation}  
In the final phase, we freeze the encoder to preserve its learned representations and train only the decoder and distillation pathways keeping the loss same as phase II. This focused optimization sharpens the model’s ability to translate distilled semantics into precise polyp segmentation masks, while minimizing overfitting and reducing computational overhead.

In summary, this strategy ensures a smooth transition from low-level feature learning to high-level semantic alignment, yielding a robust and generalizable segmentation framework.

\section{Experiments}
\subsection{Datasets}
We evaluate our proposed Polyp-DiFoM on five benchmark datasets: Kvasir-SEG \cite{jha2019kvasir}, CVC-ClinicDB \cite{bernal2015wm}, CVC-ColonDB \cite{zhao2021automatic}, ETIS \cite{silva2014toward}, and EndoScene \cite{vazquez2017benchmark}. For a fair comparison and to assess generalization performance, we strictly follow the train-test split as adopted in \cite{fan2020pranet}. Specifically, the training set consists of 1,450 images, with 900 images from Kvasir-SEG and 550 images from CVC-ClinicDB, while the testing set contains the remaining 100 and 62 images from Kvasir-SEG and CVC-ClinicDB, respectively. To further validate generalization across unseen domains, we consider 380 images from CVC-ColonDB, 196 images from ETIS, and 60 images from CVC-300 (the test set of EndoScene).

\subsection{Evaluation Metrics}
We employ six widely used metrics for polyp segmentation: IoU, Dice coefficient, Structure-measure ($S_\alpha$) \cite{fan2017structure}, Enhanced-alignment measure ($E^{\text{max}}_\phi$) \cite{fan2018enhanced}, weighted F-measure ($F^w_\beta$) \cite{margolin2014evaluate}, and Mean Absolute Error (MAE). We report the mean values of Dice and IoU, denoted as mDice and mIoU, respectively. 

\begin{table*}[ht]
\centering
\caption{Comparison of Polyp-DiFoM-enhanced baselines with SOTA methods on Kvasir-SEG and CVC-ClinicDB (seen).}
\vspace{-0.2cm}
\label{tab:combined_seen}
\resizebox{\textwidth}{!}{%
\begin{tabular}{l|c|c|c|c|c|c|c|c|c|c|c|c|c|c}
\hline
\multirow{2}{*}{Methods} & \multirow{2}{*}{Params (M)} & \multirow{2}{*}{FLOPs (G)} & \multicolumn{6}{c|}{Kvasir-SEG (Seen)} & \multicolumn{6}{c}{CVC-ClinicDB (Seen)} \\ \cmidrule(lr){4-15} 
 & & & mDice ↑ & mIoU ↑ & $F_{\beta}^{w}$ ↑ & $S_{\alpha}$ ↑ & $E_{\phi}^{\text{max}}$ ↑ & MAE ↓ & mDice ↑ & mIoU ↑ & $F_{\beta}^{w}$ ↑ & $S_{\alpha}$ ↑ & $E_{\phi}^{\text{max}}$ ↑ & MAE ↓ \\ \midrule

\rowcolor{gray!15}
\multicolumn{15}{c}{\large \textbf{Prominent State-of-the-art Methods}} \\

\rowcolor{gray!15} SANet \cite{wei2021shallow}         & 23.8 & 11.3 & 90.4 & 84.7 & 89.2 & 91.5 & 95.3 & 2.8 & 91.6 & 85.9 & 90.9 & 93.9 & 97.6 & 1.2 \\ 
\rowcolor{gray!15} MSNet \cite{zhao2021automatic}      & 27.6 & 17.0 & 90.7 & 86.2 & 89.3 & 92.2 & 94.4 & 2.8 & 92.1 & 87.9 & 91.4 & 94.1 & 97.2 & 0.8 \\
\rowcolor{gray!15} Polyp-PVT\cite{Dong2023}    & 25.1 & 10.1 & 91.7 & 86.4 & 91.1 & 92.5 & 95.6 & 2.3 & 93.7 & 88.9 & 93.6 & 94.9 & 98.5 & 0.6 \\
\rowcolor{gray!15} M$^2$SNet \cite{zhao2023m}          & 27.7 & 17.1 & 91.2 & 86.1 & 90.1 & 92.2 & 95.3 & 2.5 & 92.2 & 88.0 & 91.7 & 94.2 & 97.0 & 0.9 \\ 
\rowcolor{gray!15} PVT-Cascade \cite{rahman2023medical} & 35.2 & 32.5 & 91.1 & 86.3 & 90.6 & 91.9 & 96.1 & 2.5 & 91.9 & 87.2 & 91.8 & 93.6 & 96.9 & 1.3 \\ 
\rowcolor{gray!15} CTNet \cite{xiao2024ctnet}          & 44.2 & 32.6 & 91.7 & 86.3 & 91.0 & 92.8 & 95.9 & 2.3 & 93.6 & 88.7 & 93.4 & 95.2 & 98.3 & 0.6 \\ 
\rowcolor{gray!15} MEGANet \cite{bui2024meganet}       & 44.1 & 28.8 & 91.3 & 86.3 & 90.7 & 91.8 & 95.9 & 2.5 & 93.8 & 89.4 & 94.0 & 95.0 & 98.6 & 0.6 \\ 
\rowcolor{gray!15} CFA-Net \cite{zhou2023cross}        & 25.2 & 55.3 & 91.5 & 86.1 & 90.3 & 92.4 & 96.2 & 2.3 & 93.3 & 88.3 & 92.4 & 95.0 & 98.9 & 0.7 \\ 
\rowcolor{gray!15} SAM-Mamba \cite{dutta2025sam}       & 103.0 & 423.0 & 92.4 & 87.3 & 94.2 & 93.6 & 96.1 & 2.5 & 94.2 & 88.7 & 94.3 & 95.5 & 98.2 & 0.6 \\ \midrule

\rowcolor{cyan!10}
\multicolumn{15}{c}{\large \textbf{Lightweight Baseline Methods}} \\

\rowcolor{cyan!10} U-Net \cite{ronneberger2015u}       & 16.7 & 73.9 & 81.8 & 74.6 & 79.4 & 85.8 & 89.3 & 5.5 & 82.3 & 75.5 & 81.1 & 88.9 & 95.4 & 1.9 \\
\rowcolor{cyan!10} U-Net++ \cite{zhou2018unet++}       & 9.1 & 65.9 & 82.1 & 74.3 & 80.8 & 86.2 & 91.0 & 4.8 & 79.4 & 72.9 & 78.5 & 87.3 & 93.1 & 2.2 \\ 
\rowcolor{cyan!10} PraNet \cite{fan2020pranet}         & 30.4 & 13.1 & 89.8 & 84.0 & 88.5 & 91.5 & 94.8 & 3.0 & 89.9 & 84.9 & 89.6 & 93.6 & 97.9 & 0.9 \\ 
\rowcolor{cyan!10} EMCADNet \cite{rahman2024emcad}     & 26.7 & 21.5 & 89.4 & 83.2 & 89.2 & 96.3 & 93.5 & 3.1 & 94.4 & 89.7 & 94.3 & 99.6 & 98.7 & 0.9 \\ \midrule

\rowcolor{green!10}
\multicolumn{15}{c}{\large \textbf{Lightweight Baselines Enhanced \textcolor{red}{with Polyp-DiFoM (Ours)}}} \\

\rowcolor{green!10}
\textbf{U-Net + Ours }       & 16.9 & 74.7 & 88.5 & 78.9 & 86.4 & 95.0 & 92.0 & 3.9 & 94.9 & 90.1 & 93.6 & 98.9 & 97.8 & 0.8 \\
\rowcolor{green!5}
                    &      &      & (\textbf{\textcolor{red}{+6.7}}) & (\textbf{\textcolor{red}{+4.3}}) & (\textbf{\textcolor{red}{+7.0}}) & (\textbf{\textcolor{red}{+9.2}}) & (\textbf{\textcolor{red}{+2.7}}) & (\textcolor{blue}{-1.6}) & (\textbf{\textcolor{red}{+12.6}}) & (\textbf{\textcolor{red}{+14.6}}) & (\textbf{\textcolor{red}{+12.5}}) & (\textbf{\textcolor{red}{+10.0}}) & (\textbf{\textcolor{red}{+2.4}}) & (\textcolor{blue}{-1.1}) \\

\rowcolor{green!10}
\textbf{U-Net++ + Ours}      & 9.6  & 66.4 & 84.9 & 76.5 & 84.1 & 95.1 & 91.7 & 4.6 & 91.0 & 85.1 & 92.1 & 98.6 & 97.4 & 1.4 \\
\rowcolor{green!5}
                    &      &      & (\textbf{\textcolor{red}{+2.8}}) & (\textbf{\textcolor{red}{+2.2}}) & (\textbf{\textcolor{red}{+3.3}}) & (\textbf{\textcolor{red}{+8.9}}) & (\textbf{\textcolor{red}{+0.7}}) & (\textcolor{blue}{-0.2}) & (\textbf{\textcolor{red}{+11.6}}) & (\textbf{\textcolor{red}{+12.2}}) & (\textbf{\textcolor{red}{+13.6}}) & (\textbf{\textcolor{red}{+11.3}}) & (\textbf{\textcolor{red}{+4.3}}) & (\textcolor{blue}{-0.8}) \\

\rowcolor{green!10}
\textbf{PraNet + Ours }      & 31.4 & 13.5 & 91.2 & 84.7 & 89.2 & 96.2 & 93.4 & 2.8 & 95.3 & 91.2 & 94.9 & 99.6 & 98.9 & 0.8 \\
\rowcolor{green!5}
                    &      &      & (\textbf{\textcolor{red}{+1.4}}) & (\textbf{\textcolor{red}{+0.7}}) & (\textbf{\textcolor{red}{+0.7}}) & (\textbf{\textcolor{red}{+4.7}}) & (-1.4) & (\textcolor{blue}{-0.2}) & (\textbf{\textcolor{red}{+5.4}}) & (\textbf{\textcolor{red}{+6.3}}) & (\textbf{\textcolor{red}{+5.3}}) & (\textbf{\textcolor{red}{+6.0}}) & (\textbf{\textcolor{red}{+1.0}}) & (\textcolor{blue}{-0.1}) \\

\rowcolor{green!10}
\textbf{EMCADNet + Ours}     & 27.2 & 22.7 & 90.9 & 85.9 & 90.9 & 96.8 & 94.5 & 2.6 & 95.8 & 91.9 & 95.3 & 99.6 & 99.0 & 0.7 \\
\rowcolor{green!5}
                    &      &      & (\textbf{\textcolor{red}{+1.5}}) & (\textbf{\textcolor{red}{+2.7}}) & (\textbf{\textcolor{red}{+1.7}}) & (\textbf{\textcolor{red}{+0.5}}) & (\textbf{\textcolor{red}{+1.0}}) & (\textcolor{blue}{-0.5}) & (\textbf{\textcolor{red}{+1.4}}) & (\textbf{\textcolor{red}{+2.2}}) & (\textbf{\textcolor{red}{+1.0}}) & (\textbf{\textcolor{red}{+0.0}}) & (\textbf{\textcolor{red}{+0.3}}) & (\textcolor{blue}{-0.2}) \\
\bottomrule

\end{tabular}%
}
\vspace{-0.3cm}
\end{table*}

\begin{table*}[ht]
\centering
\caption{Comparison of Polyp-DiFoM-enhanced baselines with SOTA methods on CVC-300, CVC-ColonDB, and ETIS (unseen). }
\vspace{-0.2cm}
\label{tab:combined_unseen}
\resizebox{\textwidth}{!}{%
\begin{tabular}{l|cccccc|cccccc|cccccc}
\hline
\multirow{2}{*}{Methods} 
  & \multicolumn{6}{c|}{CVC-300 (Unseen)} 
  & \multicolumn{6}{c|}{CVC-ColonDB (Unseen)} 
  & \multicolumn{6}{c}{ETIS (Unseen)} \\ 
\cmidrule(lr){2-7}\cmidrule(lr){8-13}\cmidrule(lr){14-19}
 & mDice ↑ & mIoU ↑ & $F_{\beta}^{w}$ ↑ & $S_{\alpha}$ ↑ & $E_{\phi}^{\max}$ ↑ & MAE ↓ 
 & mDice ↑ & mIoU ↑ & $F_{\beta}^{w}$ ↑ & $S_{\alpha}$ ↑ & $E_{\phi}^{\max}$ ↑ & MAE ↓ 
 & mDice ↑ & mIoU ↑ & $F_{\beta}^{w}$ ↑ & $S_{\alpha}$ ↑ & $E_{\phi}^{\max}$ ↑ & MAE ↓ \\ 
\midrule

\rowcolor{gray!15}
\multicolumn{19}{c}{\large \textbf{Prominent State-of-the-art Methods}} \\

\rowcolor{gray!15} SANet \cite{wei2021shallow}       
  & 88.8 & 81.5 & 85.9 & 92.8 & 97.2 & 0.8  
  & 75.3 & 67.0 & 72.6 & 83.7 & 87.8 & 4.3  
  & 75.0 & 65.4 & 68.5 & 84.9 & 89.7 & 1.5 \\

\rowcolor{gray!15} MSNet \cite{zhao2021automatic}    
  & 86.9 & 80.7 & 84.9 & 92.5 & 94.3 & 1.0  
  & 75.5 & 67.8 & 73.7 & 83.6 & 88.3 & 4.1  
  & 71.9 & 66.4 & 67.8 & 84.0 & 83.0 & 2.0 \\

\rowcolor{gray!15} Polyp-PVT \cite{Dong2023}  
  & 90.0 & 83.3 & 88.4 & 93.5 & 97.3 & 0.7  
  & 80.8 & 72.7 & 79.5 & 86.5 & 91.3 & 3.1  
  & 78.7 & 70.6 & 75.0 & 87.1 & 90.6 & 1.3 \\

\rowcolor{gray!15} M\textsuperscript{2}SNet \cite{zhao2023m}  
  & 90.3 & 84.2 & 88.1 & 93.9 & 96.5 & 0.9  
  & 75.8 & 68.5 & 73.7 & 84.2 & 86.9 & 3.8  
  & 74.9 & 67.8 & 71.2 & 84.6 & 87.2 & 1.7 \\

\rowcolor{gray!15} PVT-Cascade \cite{rahman2023medical}  
  & 89.2 & 82.4 & 87.3 & 93.2 & 95.9 & 0.9  
  & 78.1 & 71.0 & 77.9 & 85.5 & 89.6 & 3.1  
  & 78.6 & 71.2 & 75.9 & 87.2 & 89.6 & 1.3 \\

\rowcolor{gray!15} CTNet \cite{xiao2024ctnet}         
  & 90.8 & 84.4 & 89.4 & 97.5 & 97.5 & 0.6  
  & 81.3 & 73.4 & 80.1 & 87.4 & 91.5 & 2.7  
  & 81.0 & 73.4 & 77.6 & 88.6 & 91.3 & 1.4 \\

\rowcolor{gray!15} MEGANet \cite{bui2024meganet}      
  & 89.9 & 83.4 & 88.2 & 93.5 & 96.9 & 0.7  
  & 79.3 & 71.4 & 77.9 & 85.4 & 89.5 & 4.0  
  & 73.9 & 66.5 & 70.2 & 83.6 & 85.8 & 3.7 \\

\rowcolor{gray!15} CFA-Net \cite{zhou2023cross}       
  & 89.3 & 82.7 & 93.8 & 87.5 & 97.8 & 0.8  
  & 74.3 & 66.5 & 72.8 & 83.5 & 89.8 & 3.9  
  & 73.2 & 65.5 & 69.3 & 84.5 & 89.2 & 1.4 \\

\rowcolor{gray!15} SAM-Mamba \cite{dutta2025sam}      
  & 92.0 & 86.1 & 88.8 & 94.6 & 98.1 & 0.6  
  & 85.3 & 77.1 & 85.6 & 89.8 & 93.3 & 1.7  
  & 84.8 & 78.2 & 85.5 & 91.6 & 93.3 & 1.0 \\ 
\midrule

\rowcolor{cyan!10}
\multicolumn{19}{c}{\large \textbf{Lightweight Baseline Methods}} \\

\rowcolor{cyan!10} U-Net \cite{ronneberger2015u}      
  & 71.0 & 62.7 & 68.4 & 84.3 & 87.6 & 2.2  
  & 51.2 & 44.4 & 49.8 & 71.2 & 77.6 & 6.1  
  & 39.8 & 33.5 & 36.6 & 68.4 & 74.0 & 3.6 \\

\rowcolor{cyan!10} U-Net++ \cite{zhou2018unet++}      
  & 70.7 & 62.4 & 68.7 & 83.9 & 89.8 & 1.8  
  & 48.3 & 41.0 & 46.7 & 69.1 & 76.0 & 6.4  
  & 40.1 & 34.4 & 39.0 & 68.3 & 77.6 & 3.5 \\

\rowcolor{cyan!10} PraNet \cite{fan2020pranet}        
  & 87.1 & 79.7 & 84.3 & 92.5 & 97.2 & 1.0  
  & 70.9 & 64.0 & 69.6 & 81.9 & 86.9 & 4.5  
  & 62.8 & 56.7 & 60.0 & 79.4 & 84.1 & 3.1 \\

\rowcolor{cyan!10} EMCADNet \cite{rahman2024emcad}   
  & 87.8 & 81.9 & 88.6 & 98.3 & 96.5 & 0.7  
  & 76.0 & 68.0 & 75.7 & 92.1 & 88.3 & 3.8  
  & 74.4 & 65.2 & 74.0 & 91.9 & 87.8 & 2.0 \\ 
\midrule

\rowcolor{green!10}
\multicolumn{19}{c}{\large \textbf{Lightweight Baselines Enhanced \textcolor{red}{with Polyp-DiFoM (Ours)}}} \\

\rowcolor{green!10} \textbf{U-Net + Ours}       
  & 86.2 & 73.3 & 77.3 & 91.2 & 87.7 & 1.2  
  & 67.3 & 57.8 & 60.8 & 81.6 & 77.0 & 4.8  
  & 55.1 & 43.7 & 48.2 & 77.1 & 69.4 & 2.4 \\
\rowcolor{green!5}
  & (\textbf{\textcolor{red}{+15.2}}) & (\textbf{\textcolor{red}{+10.6}}) & (\textbf{\textcolor{red}{+8.9}}) & (\textbf{\textcolor{red}{+6.9}}) & (\textbf{\textcolor{red}{+0.1}}) & (\textcolor{blue}{-1.0}) 
    & (\textbf{\textcolor{red}{+16.1}}) & (\textbf{\textcolor{red}{+13.4}}) & (\textbf{\textcolor{red}{+11.0}}) & (\textbf{\textcolor{red}{+10.4}}) & (-0.6) & (\textcolor{blue}{-1.3}) 
    & (\textbf{\textcolor{red}{+15.3}}) & (\textbf{\textcolor{red}{+10.2}}) & (\textbf{\textcolor{red}{+11.6}}) & (\textbf{\textcolor{red}{+8.7}}) & (-4.6) & (\textcolor{blue}{-1.2}) \\

\rowcolor{green!10} \textbf{U-Net++ + Ours}      
  & 84.8 & 76.4 & 78.5 & 93.8 & 91.3 & 1.3  
  & 69.1 & 58.8 & 64.2 & 86.4 & 81.5 & 4.6  
  & 53.6 & 43.5 & 49.2 & 80.8 & 72.5 & 3.0 \\
\rowcolor{green!5}
  & (\textbf{\textcolor{red}{+14.1}}) & (\textbf{\textcolor{red}{+14.0}}) & (\textbf{\textcolor{red}{+9.8}}) & (\textbf{\textcolor{red}{+9.9}}) & (\textbf{\textcolor{red}{+1.5}}) & (\textcolor{blue}{-0.5}) 
    & (\textbf{\textcolor{red}{+17.8}}) & (\textbf{\textcolor{red}{+17.5}}) & (\textbf{\textcolor{red}{+17.5}}) & (\textbf{\textcolor{red}{+17.3}}) & (\textbf{\textcolor{red}{+5.5}}) & (\textcolor{blue}{-1.8}) 
    & (\textbf{\textcolor{red}{+13.5}}) & (\textbf{\textcolor{red}{+9.1}}) & (\textbf{\textcolor{red}{+10.2}}) & (\textbf{\textcolor{red}{+12.5}}) & (-5.1) & (\textcolor{blue}{-0.5}) \\

\rowcolor{green!10} \textbf{PraNet + Ours}       
  & 90.0 & 83.0 & 89.7 & 97.3 & 98.6 & 0.6  
  & 75.3 & 67.2 & 73.4 & 89.1 & 85.8 & 3.8  
  & 73.0 & 62.3 & 67.2 & 89.1 & 84.0 & 1.6 \\
\rowcolor{green!5}
  & (\textbf{\textcolor{red}{+2.9}}) & (\textbf{\textcolor{red}{+3.3}}) & (\textbf{\textcolor{red}{+5.4}}) & (\textbf{\textcolor{red}{+4.8}}) & (\textbf{\textcolor{red}{+1.4}}) & (\textcolor{blue}{-0.4}) 
    & (\textbf{\textcolor{red}{+4.4}}) & (\textbf{\textcolor{red}{+3.2}}) & (\textbf{\textcolor{red}{+3.8}}) & (\textbf{\textcolor{red}{+7.2}}) & (-1.1) & (\textcolor{blue}{-0.7}) 
    & (\textbf{\textcolor{red}{+10.2}}) & (\textbf{\textcolor{red}{+5.6}}) & (\textbf{\textcolor{red}{+7.2}}) & (\textbf{\textcolor{red}{+9.7}}) & (-0.1) & (\textcolor{blue}{-1.5}) \\

\rowcolor{green!10} \textbf{EMCADNet + Ours}     
  & 88.9 & 80.5 & 86.7 & 97.4 & 94.4 & 1.0  
  & 77.6 & 69.8 & 77.5 & 92.7 & 88.7 & 3.9  
  & 75.9 & 68.0 & 74.6 & 91.9 & 88.3 & 2.5 \\
\rowcolor{green!5}
  & (\textbf{\textcolor{red}{+1.1}}) & (-1.4) & (-1.9) & (-0.9) & (-2.1) & (+0.3) 
    & (\textbf{\textcolor{red}{+1.6}}) & (\textbf{\textcolor{red}{+1.8}}) & (\textbf{\textcolor{red}{+1.8}}) & (\textbf{\textcolor{red}{+0.6}}) & (\textbf{\textcolor{red}{+0.4}}) & (+0.1) 
    & (\textbf{\textcolor{red}{+1.5}}) & (\textbf{\textcolor{red}{+2.8}}) & (\textbf{\textcolor{red}{+0.6}}) & (0.0) & ( \textbf{\textcolor{red}{+0.5}}) & (\textbf{+0.5}) \\  

\bottomrule
\end{tabular}%
}
\vspace{-0.5cm}
\end{table*}

\subsection{Implementation Details}
Our model has been developed using the PyTorch framework and is trained on a single Tesla V100 GPU with 32 GB of memory. We resize all images 352 × 352 pixels. Data augmentation involves random flipping and multi-scaling with scales of {0.75, 1.0, and 1.25}.
Adam optimizer with a learning rate of 1 × $10^{-4}$ is used to train the model. The model is trained for 120 epochs using a three-phase training strategy. In \textbf{Phase I}, the model is trained for 40 epochs using combined dice and binary cross entropy loss. \textbf{Phase II} is from epochs 40 to 80, in which we include a distillation loss alongside the segmentation loss to guide the student model. Finally, during epochs 80 to 120 (\textbf{Phase III}), the encoder is frozen while the decoder and task-specific layers are trained with same loss used in phase-II. This progressive training helps balance segmentation performance and knowledge transfer.

\begin{figure*}[t]
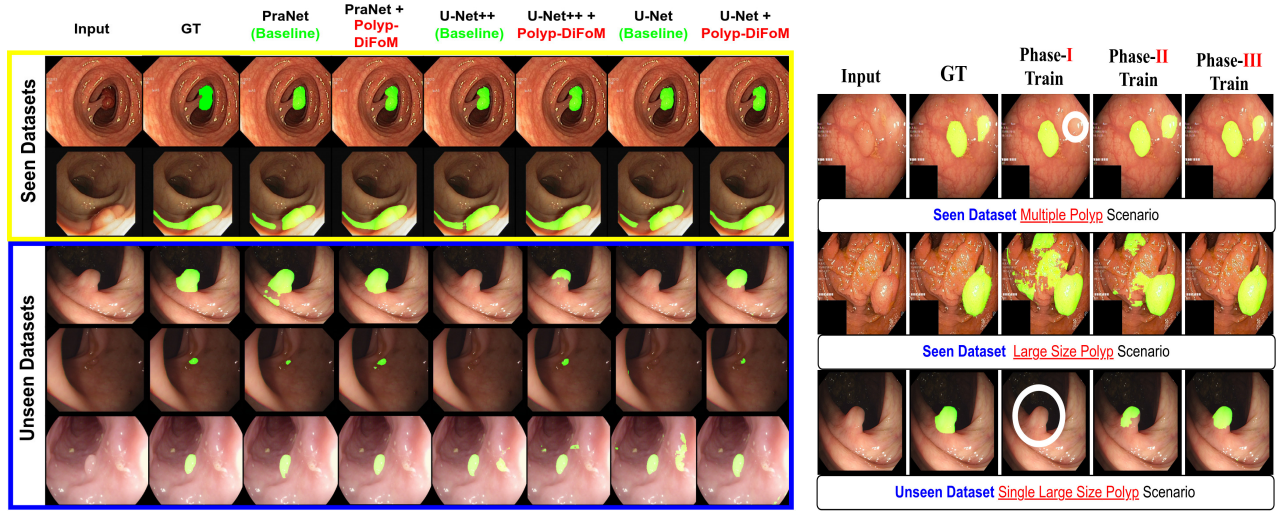

  \centering
  \begin{subfigure}[t]{0.6\linewidth}
    \centering
    \includegraphics[width=\linewidth,height=0.3\textheight]{Images/QA1Polyp.pdf}
    \caption{Qualitative comparison on seen datasets (Kvasir-SEG and CVC-ClinicDB), and on unseen datasets (CVC-300, CVC-ColonDB and ETIS), demonstrating the model’s capability to accurately segment polyps of varying sizes, colors and textures.}
    \vspace{-0.1cm}
    \label{fig:qa1}
  \end{subfigure}
  \hspace{0.1cm}
  \begin{subfigure}[t]{0.35\linewidth}
    \centering
    \includegraphics[width=\linewidth,height=0.28\textheight]{Images/QA2Polyp.pdf}
    \caption{Effect of multi-objective training strategy considering U-Net as baseline. Here, the white circle denotes the undetected region.}
    \label{fig:qa2}
    \vspace{-0.1cm}
  \end{subfigure}
  
  \caption{Qualitative results comparison across all five datasets with an illustration of phase-wise learning progression}
  \label{fig:qualitative}
\vspace{-0.5cm}
  
\end{figure*}

\section{State-of-the-art Comparison}
To experimentally validate the robustness of our Polyp-DiFoM, we select four lightweight baseline methods, such as U-Net \cite{ronneberger2015u}, U-Net++ \cite{zhou2018unet++}, PraNet \cite{fan2020pranet}, and EMCADNet \cite{rahman2024emcad}. \\
\textbf{\underline{Seen Dataset Precision :}}
We compare Polyp-DiFoM against baseline architectures for seen datasets Kvasir-SEG and CVC-ClinicDB. As summarized in Table \ref{tab:combined_seen}, our method achieves consistent improvements over all baseline models while maintaining almost the same parameter count and FLOPs. For instance, on CVC-ClinicDB, baseline U-Net Dice improves from 82.3 to 94.9 and IoU from 75.5 to 90.1, while reducing MAE from 1.9 to 0.8, with only a marginal increase in parameters (16.7M to 16.9M). Similarly, PraNet surpasses its baseline by +1.4\% Dice and +0.7\% IoU on Kvasir-SEG, and achieves 95.3\%  Dice on CVC-ClinicDB, approaching the performance of heavier models like SAM-Mamba, but with significantly lower parameters. Although, PraNet and U-Net++ show modest numerical improvements for Kvasir-SEG, Figure~\ref{fig:qa1} demonstrates clear qualitative improvements where our model accurately captures low-contrast and boundary-adjacent polyps than their respective baselines. As model complexity increases, the benefit of distillation becomes marginal since complex architectures already possess rich representational power. For example, in EMCADNet, the performance gains are relatively smaller, which is expected given it already leverages rich encoder–decoder feature extraction. 
\newline
\textbf{\underline{Cross Dataset Generalization Robustness :}}
We also compare Polyp-DiFoM on three unseen datasets: CVC-300, CVC-ColonDB, and ETIS. The results presented in Table \ref{tab:combined_unseen} clearly demonstrate that our Polyp-DiFoM consistently improves cross-dataset performance. For instance, on CVC-300, Dice score of U-Net baseline improves from 71.0\% to 86.2\%, while U-Net++ IoU increases from 62.4 to 76.4\%. Similarly, PraNet achieves 90.0\% Dice on CVC-300 and 73.0\% Dice on ETIS, significantly outperforming its baseline counterpart (87.1\% and 62.8\%). EMCADNet also achieves competitive results across unseen datasets, with notable gains in Dice and $F_{\beta}^{w}$ while maintaining similar complexity. The improved generalization arises from the distillation process, where the teacher model transfers rich semantic priors, enabling baselines to overcome dataset-specific biases and learn more transferable features. As a result, Polyp-DiFoM is better at handling appearance variations, illumination changes, and complex polyp boundaries in unseen datasets.
\newline
\textbf{\underline{Tradeoff Analysis and Discussion :}}
The results clearly show that Polyp-DiFoM achieves competitive performance compared to recent state-of-the-art methods, despite being significantly lighter. Models such as CTNet \cite{xiao2024ctnet} and MEGANet \cite{bui2024meganet} demand more than twice the parameters, while SAM-Mamba \cite{dutta2025sam} is over 9$\times$ larger, yet our Polyp-DiFoM variant consistently achieves results that are on par with these models in both the seen and unseen categories. The improvements offered by Polyp-DiFoM underscore the effectiveness of distilling foundation models to enhance lightweight baselines with minimal additional computational cost. More importantly, the gains on unseen datasets highlight improved generalization, making Polyp-DiFoM not only efficient but also robust across diverse clinical scenarios.

\section{Qualitative Comparison}
To further evaluate the effectiveness of our approach, we conduct qualitative comparisons between baseline models and their enhanced counterparts for both seen and unseen datasets. As shown in Figure \ref{fig:qa1}, baseline architectures often struggle with challenging scenarios such as weak boundaries, small-scale polyps, or regions with low contrast to their surroundings, where segmentation masks tend to either miss subtle structures or include spurious predictions along edges. Polyp-DiFoM variants consistently produce sharper boundaries and more complete polyp regions. 

For instance, in U-Net++, the baseline predictions occasionally over-segment background textures, whereas the Polyp-DiFoM demonstrates improved discrimination between true polyp regions and surrounding tissue. Due to poor generalization, baselines often fail to detect polyps; however, Polyp-DiFoM generates more precise and reliable segmentation masks, thanks to its improved generalizability. These observations demonstrate that our framework not only enhances quantitative performance but also yields qualitatively superior results across diverse datasets.

Further, Figure~\ref{fig:qa2} highlights the effect of our multi-objective training strategy. In cases with multiple polyp regions, early training using only the segmentation loss tends to capture coarse structures but often fails to detect secondary regions due to the absence of global contextual cues. During the middle phase, when distillation is introduced, the teacher guides the baseline to discover finer and missed regions, and the predictions improve. Finally, encoder freezing prevents overfitting and preserves distilled knowledge, resulting in more accurate detection with sharper boundaries.
In another case of an unseen single large size polyp, the polyp is missed entirely in the early stage, begins to appear with distillation in the middle phase, and is segmented most accurately at the end. However, the same sample, when trained with a standard baseline using only segmentation loss, remains entirely undetected throughout the training.

\section{Ablation Study}
\textbf{\underline{Effect of Foundation Models: }}
To verify the significance of using multiple foundational models, we progressively incorporate them into the baselines, as shown in Table~\ref{tab:ABL1}. These models include CLIP, DINOv2, OneFormer, Mask2Former, OWL-ViT \cite{minderer2022simple} and I-JEPA \cite{assran2023self}. While SAM alone produces results closer to the baseline, SAM with CLIP leads to a performance drop likely because CLIP’s representations are not well-optimized for dense, pixel-level tasks such as segmentation. However, SAM with DINOv2 improves performance over SAM, indicating the benefit of incorporating vision-based features. Adding OWL-ViT and I-JEPA with SAM + DINOv2 further leads to performance drops, suggesting that not all foundation models provide synergistic representations. The combination of SAM + DINOv2 + OneFormer achieves significant improvements, and incorporating Mask2Former as a fourth FM produces the overall best performance. However, models with strong representational capacity do not gain as much from increasing the number of foundation models. This is evident in the case of EMCADNet, where the configuration with three foundation models actually outperforms the configuration using all four.
\newline
\textbf{\underline{Effect of Frequency-Domain Encoding:}} In Table~\ref{tab:loss_freq_ablation}, we compare the performance with FFT and  Discrete Cosine Transform  (DCT) based encoding. FFT, helps to effectively separate high- and low-frequency components, allowing the baseline to learn both fine-grained details and global context from the teacher model. In contrast, DCT-based encoding does not yield significant gains, likely because DCT is less effective at capturing localized variations and complex structural details in medical images. 
\newline
\textbf{\underline{Effect of Distillation Loss:}}
We evaluate the effectiveness of different combinations of distillation loss for transferring knowledge from the teacher model to the baseline network. Specifically, we consider three configurations: MSE, MSE + KL divergence, and MSE + MAE combinations. Table~\ref{tab:loss_freq_ablation} shows that MSE alone yields the best performance which ensures local feature alignment and focuses on region overlap while combining with Dice loss, resulting in complementary optimization. In contrast, adding KL divergence or MAE introduces less aligned supervision, leading to suboptimal results.
It is worth noting that the correct combination of foundational models,  frequency-domain encoding, and distillation loss is critical to maximize the performance of Polyp-DiFoM.

\begin{table}[t]
\small
\centering
\caption{Ablation study to show the effectiveness of foundation models in four baselines.}
\vspace{-0.1cm}
\resizebox{\columnwidth}{!}{
\begin{tabular}{cccc|cc|ccc}
    \toprule
    &&&& \multicolumn{2}{c|}{\bf \small Seen}&\multicolumn{3}{c}{\bf   \small Unseen}\\
    
    \ \textbf{SAM} & \textbf{DINOv2} & \textbf{OneFormer} & \textbf{Mask2Former} &\bf   \small Kavsir & \bf   \small CVC-ClinicDB& \bf   \small CVC-300 &   \bf \small CVC-ColonDB& \bf   \small ETIS \\\midrule
    
    \rowcolor{yellow!15}
    \multicolumn{9}{c}{\textbf{\textcolor{red}{U-Net}}} \\
    
    \checkmark & - & - & - &  81.7 & 87.6 & 72.7 & 57.6  & 43.9 \\
    \checkmark & \checkmark & - & - & 85.4  & 92.0 & 80.4  & \textbf{67.9} & 51.9  \\
    \checkmark & \checkmark & \checkmark & - & 86.6 & 93.9 & 82.3 & 68.3 & \textbf{54.3} \\
    \checkmark & \checkmark & \checkmark & \checkmark & \textbf{88.5} & \textbf{94.9} & \textbf{86.2} & 67.3 & \textbf{55.1} \\ \midrule

    \rowcolor{yellow!15}
    \multicolumn{9}{c}{\textbf{\textcolor{red}{U-Net++ }}} \\
    
    \checkmark & - & - & -  & 81.7 & 85.6 & 71.9 & 56.3 & 42.2 \\
    \checkmark & \checkmark & - & - & 81.8  & 87.8  & 76.5  & 60.9  & 48.2 \\
    \checkmark & \checkmark & \checkmark & - & 84.7 & 89.5 & 77.8 & 67.7 & 51.8 \\
    \checkmark & \checkmark & \checkmark & \checkmark & \textbf{84.9} & \textbf{91.0} & \textbf{84.8} & \textbf{69.1} & \textbf{53.6} \\ \midrule

    \rowcolor{yellow!15}
    \multicolumn{9}{c}{\textbf{\textcolor{red}{PraNet }}} \\
    
    \checkmark & - & - & - &  89.9 & 90.8 & 86.3 & 71.7 & 63.8  \\
    \checkmark & \checkmark & - & - & 89.9  & 93.4  & 86.7  & 73.9  & 70.7   \\
    \checkmark & \checkmark & \checkmark & -  & 90.9 & 94.2 & 87.4 & 74.0 & \textbf{71.5} \\
    \checkmark & \checkmark & \checkmark & \checkmark  & \textbf{91.2} & \textbf{95.3} & \textbf{90.0} & \textbf{75.3} & \textbf{73.0} \\ \midrule

    \rowcolor{yellow!15}
    \multicolumn{9}{c}{\textbf{\textcolor{red}{EMCADNet }}} \\
    
    \checkmark & - &  - & - & 89.5 & 94.4  & 86.6 & 74.4 & 73.7  \\
    \checkmark & \checkmark & - & - & 90.8  & 94.5  & 88.8 & 75.2  & 75.4 \\
    \checkmark & \checkmark & \checkmark & - & \textbf{91.3} & \textbf{95.9} & \textbf{89.3} & \textbf{78.3}  & \textbf{76.7}    \\
    \checkmark & \checkmark & \checkmark & \checkmark & 90.9 & 95.8  & 88.9  & 77.6 & 75.9 \\ 




    \bottomrule
\end{tabular}
}

\label{tab:ABL1}
\end{table}

\renewcommand{\arraystretch}{0.85}
\begin{table}[t]
\centering
\caption{Ablation on distillation losses and frequency-based encoding for U-Net + Polyp-DiFoM. }
\label{tab:loss_freq_ablation}
\resizebox{\columnwidth}{!}{%
\begin{tabular}{c c c c c | c c}
\toprule
\textbf{MSE} & \textbf{KL} & \textbf{MAE} & \textbf{FFT} & \textbf{DCT} & \bf \small Kvasir ↑ & \bf \small CVC-ClinicDB ↑ \\ \midrule
    
    \rowcolor{yellow!15}
    \multicolumn{7}{c}{\textbf{\textcolor{red}{Distillation Loss}}} \\

    \checkmark & - & \checkmark & \checkmark & -  & 69.4 & 77.8 \\
    \checkmark & \checkmark & - & \checkmark & - & 74.7 & 82.5  \\
    \checkmark & - & - & \checkmark & - & \textbf{88.5} & \textbf{94.9} \\
 \midrule
    \rowcolor{yellow!15}
    \multicolumn{7}{c}{\textbf{\textcolor{red}{Frequency Encoding}}} \\

    \checkmark & - & - & - & \checkmark & 72.8 & 83.8  \\

\bottomrule
\end{tabular}%
}
\vspace{-0.4cm}
\end{table}

\section{Conclusion}
This paper presents Polyp-DiFoM, a novel distillation framework that transfers the rich representational power of large-scale foundation models into lightweight segmentation architectures. By infusing semantic priors from foundational models, Polyp-DiFoM enhances the accuracy and efficiency of lightweight baselines significantly. Extensive experiments across five benchmark datasets demonstrate that Polyp-DiFoM attains performance on par with recent state-of-the-art methods, while having substantially fewer parameters and significantly reduced computational overhead. This highlights the effectiveness of Polyp-DiFoM in real-time, resource-constrained clinical deployment where the use of complex and computationally heavy models is often impractical.
\paragraph{Acknowledgement:} This work is supported by the Anusandhan National Research Foundation (ANRF), Government of India, under project number CRG/2023/007397. D. Jha is supported by the University of South Dakota.
{
    \small
    \bibliographystyle{IEEEtran}
    \bibliography{main}
}

\end{document}